\documentclass[acmsmall,screen]{acmart}
\pdfoutput=1
\usepackage{graphicx}       
\usepackage{pgfplots}       
\usepackage{pgf-pie}        
\usepackage{float}          
\usepackage{caption}        
\usepackage{subcaption}     
\usepackage{booktabs}       
\usepackage{tabularx}       
\usepackage{longtable}      
\usepackage{lscape}         
\usepackage{makecell}       
\usepackage{array}          

\AtBeginDocument{%
  \providecommand\BibTeX{{%
    Bib\TeX}}}

\setcopyright{acmlicensed}
\copyrightyear{2024}
\acmYear{2024}
\acmDOI{XXXXXXX.XXXXXXX}

\acmJournal{JACM}
\acmVolume{XX}
\acmNumber{X}
\acmArticle{XXX}
\acmMonth{2}

\begin{document}

\title{Seeing the Intangible: Survey of Image Classification into High-Level and Abstract Categories}


\author{Delfina Sol Martinez Pandiani}
\email{dsmp@cwi.nl}
\orcid{0000-0003-2392-6300}
\affiliation{%
  \institution{University of Bologna}
  \streetaddress{Department of Computer Science and Engineering (DISI)}
  \city{Bologna}
  \country{Italy}
}
\affiliation{%
  \institution{Centrum Wiskunde en Informatica}
  \streetaddress{Human-Centered Data Analytics}
  \city{Amsterdam}
  \country{The Netherlands}
}

\author{Valentina Presutti}
\affiliation{%
  \institution{University of Bologna}
  \streetaddress{Department of Modern Languages, Literatures and Cultures (LILEC)}
  \city{Bologna}
  \country{Italy}
}

\renewcommand{\shortauthors}{Martinez Pandiani and Presutti}


\begin{abstract}

The field of Computer Vision (CV) is increasingly shifting towards ``high-level'' visual sensemaking tasks, yet the exact nature of these tasks remains unclear and tacit. This survey paper addresses this ambiguity by systematically reviewing research on high-level visual understanding, focusing particularly on Abstract Concepts (ACs) in automatic image classification. Our survey contributes in three main ways: Firstly, it clarifies the tacit understanding of high-level semantics in CV through a multidisciplinary analysis, and categorization into distinct clusters, including commonsense, emotional, aesthetic, and inductive interpretative semantics. Secondly, it identifies and categorizes computer vision tasks associated with high-level visual sensemaking, offering insights into the diverse research areas within this domain. Lastly, it examines how abstract concepts such as values and ideologies are handled in CV, revealing challenges and opportunities in AC-based image classification. Notably, our survey of AC image classification tasks highlights persistent challenges, such as the limited efficacy of massive datasets and the importance of integrating supplementary information and mid-level features. We emphasize the growing relevance of hybrid AI systems in addressing the multifaceted nature of AC image classification tasks. Overall, this survey enhances our understanding of high-level visual reasoning in CV and lays the groundwork for future research endeavors.
\end{abstract}



\maketitle

\section{Introduction}
\label{sec:intro}

Visual imagery has historically been a potent medium for conveying both abstract and concrete ideas, a significance evident in the vast amount of images shared daily on social media \cite{edwards2014were}. This surge in visual content has fueled extensive research in Computer Vision (CV), primarily aimed at automating the indexing, retrieval, and management of visual data, with applications spanning disciplines like sociology, media studies, and psychology \cite{joo2014visualpersuasion, arnold2019distant}. CV's data-driven approach, treating images as data, has been pivotal, facilitated further by the recent deep learning (DL) paradigm shift, leading to significant achievements in tasks such as image classification, object detection, and image generation \cite{bagi2020deep}. The remarkable success of the Deep Learning (DL) paradigm in Computer Vision (CV) has led to more intricate demands, including the need for tools capable of replicating human-like perception at a "high semantic level" \cite{hussain2017automaticunderstanding}. This includes using CV to classify images based on high-level notions, known as Abstract Concepts (ACs), which have proven instrumental in various tasks such as emotion classification \cite{cao2018emotionalmodelling, mohammad2018wikiartemotions}, political affiliation detection \cite{joo2014visualpersuasion}, beauty assessment \cite{gray2010predictingfacial}, and personality trait inference \cite{segalin2017socialprofiling}, all accomplished through raw visual data. 

However, explicit definitions of high-level visual semantics, particularly ACs, in machine vision are sparse. This lack of clarity, combined with the historical emphasis on physical object detection grounded in low-level feature analysis, often results in less impressive results in high-level semantic tasks compared to concrete object classes \cite{borghi2014wordssocial}. Additionally, these tasks are influenced by cultural contexts and human biases in perception, which redefine the depth of knowledge and understanding expected from CV models. Our survey systematically reviews CV studies addressing the challenge of automatically classifying visual data based on high-level semantic units. We clarify what constitutes ``abstract" or ``high-level" semantics in the context of an image and identify CV tasks and automatic detection approaches related to these semantics. Focusing on abstract concept-based image classification (AC image classification), particularly in still images, we conduct a comprehensive overview of the state of the art. This includes:

\begin{enumerate}
\item \textbf{High-Level Semantic Units}: Identification and clustering of high-level semantic units, integrating insights from cognitive science, visual studies, art history, and computer science.
\item \textbf{High-Level CV Tasks}: Surveying of the CV landscape to identify and cluster tasks associated with high-level visual sensemaking, while examining common methodologies and datasets.
\item \textbf{AC Image Classification}: We conduct a detailed review of works dealing explicitly with AC image classification in still images.
\end{enumerate}

This work is structured as follows. Section \ref{section:survey_tip} provides an interdisciplinary examination and characterization of what constitutes ``full'' or ``high-level'' semantics in human visual understanding. In Section \ref{section:survey_method}, the methodology employed to identify works related to the high-level semantics in the CV field is described. Section \ref{section:survey_ahlvu} surveys and categorizes CV tasks and works associated with high-level visual understanding, facilitating the discovery of implicit CV research addressing ACs. In Section \ref{section:survey_ac_ic} we perform a thorough survey of CV-based works that research tasks analogous to AC image classification. Section \ref{section:survey_datasets} presents datasets potentially relevant to the AC image classification task. The implications and contributions of the survey are discussed in Section \ref{section:survey_discussion}. Ultimately, Section \ref{section:survey_conclusion} provides concluding remarks. More details are available and documented in a specialized GitHub repository.\footnote{\url{https://github.com/delfimpandiani/seeing_the_intangible}. Access date: February 2024.}

\section{Defining High-Level Visual Semantics}
\label{section:survey_tip}

\subsection{Three-Tiered Semantics}

\begin{figure}[ht]
    \begin{minipage}{\textwidth}
        \centering
        \includegraphics[width=.8\linewidth]{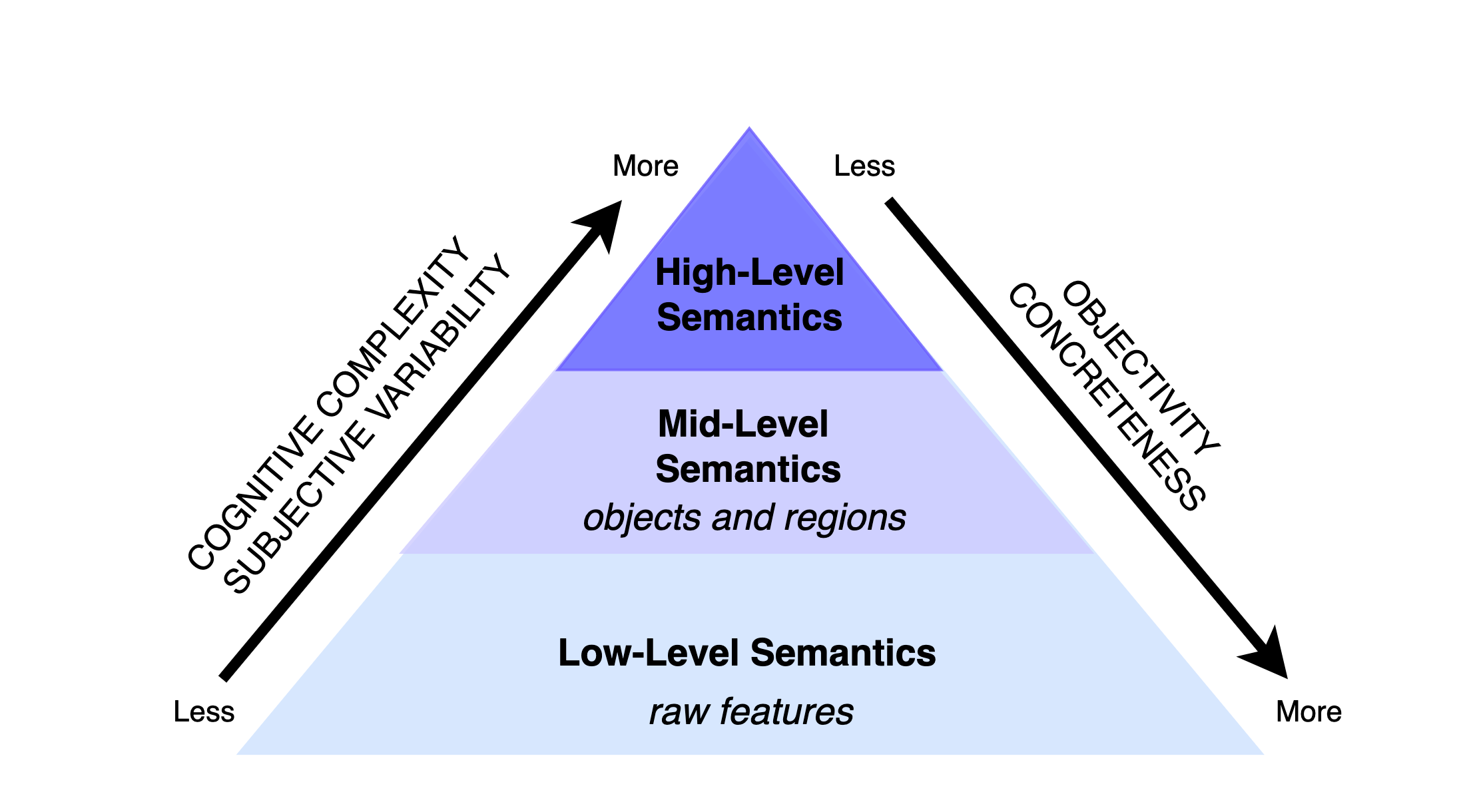}
        \caption[The three tiers of the visual semantics hierarchy.]{The three tiers of the visual semantics hierarchy. Visual understanding is often depicted as a multi-layered process, revealing three distinct levels of semantics. The low-level involves raw or elemental features, while the mid-level encompasses individual objects, persons, and regions. In contrast, the high-level remains less defined and explored.
        \label{fig:three-levels}}
    \end{minipage}
\end{figure}

The concept that the perception and interpretation of visual meaning involve a multi-layered process is a shared perspective across various domains and applications, including cognitive science, CV, content-based image retrieval (CBIR), and visual studies. This multi-layered nature was emphasized in the seminal paper by Hare et al. (2006) \cite{hare2006mind}, which discussed Smeulders' idea of the ``semantic gap" in CV \cite{smeulders2000content}. This paper also highlighted the common practice of referring to different strata of meaning within images, a concept that has been pivotal in CBIR. We delved deeper into several of these multi-layered approaches, drawing insights from works by Panofsky (1955) \cite{panofsky1955meaning}, Shatford (1986) \cite{shatford1986analyzingsubjecta}, Greisdorf and O’Connor (2002) \cite{greisdorf2002modellingwhat}, Eakins (2000) \cite{eakins2000retrieval}, Jorgensen (2003) \cite{jorgensen2003imageretrieval}, Hare et al. (2006) \cite{hare2006mind}, and Aditya et al. (2019) \cite{aditya2019integrating}. This exploration revealed a general analogy wherein three semantic tiers are used to delineate the human visual understanding process: a ``low-level," a ``mid-level," and an ``upper-" or ``high-level" tier, corresponding to increasing complexity, variability, and subjectivity (see Figure \ref{fig:three-levels}). Most of these approaches represented these layers using a pyramid analogy to illustrate a hierarchical structure. Via a thorough analysis of the semantic elements assigned to each of the layers by each of the foundational works, we noted that there was a consensus in identifying and agreeing upon semantic units within both the low- and mid-level layers. However, this consensus did not extend to the topmost layer. 

At the base, the ``low-level" layer (depicted in light blue in Figure \ref{fig:three-levels}) encompasses raw or primitive features such as regions, edges, textures, colors, shapes, and textures. Moving up to the ``mid-level" layer (depicted in light purple in Figure \ref{fig:three-levels}), this tier accommodates semantic entities like objects, persons, regions, and places. Much of CV research has centered on this layer, emphasizing object recognition and image segmentation. In contrast, the ``high-level" layer of semantics (depicted in dark purple in Figure \ref{fig:three-levels}) remains less detailed and subject to less consensus. This topmost tier, often associated with the concept of ``full semantics," lacks an explicit and consistent definition, and characterization of what types of semantic units belong in it. Instead, there appears to be a \textit{tacit} shared understanding of the kinds of content that may reside or be conceived within this layer. In our analysis, this layer emerged as both elusive and significant, akin to the ``tip of an iceberg" regarding visual semantics, motivating our efforts to define it more precisely.

\subsection{Tip of the Iceberg: Upper Visual Semantics}

Images may be sought ``on the basis of their holistic content or message, as opposed to the information embedded within them by dint of their depiction of certain features” \cite[p. 39]{enser1999visualimage}. Most work that attempts to name and characterize where and how such holistic content arises thus moves in a layered way further away from raw or primitive features, to arrive to the “highest” tier of the semantic pyramid, referred to with different names: iconological layer \cite{panofsky1955meaning}, higher level of understanding \cite{jorgensen2003imageretrieval}, abstract content  \cite{shatford1986analyzingsubjecta}, abstract attributes \cite{eakins2000retrieval}, subjective beliefs \cite{greisdorf2002modellingwhat}, higher level semantics \cite{aditya2019integrating}, or full semantics \cite{hare2006mind}.

\begin{figure}[!h]
    \begin{minipage}{\textwidth}
        \centering
        \includegraphics[width=\linewidth]{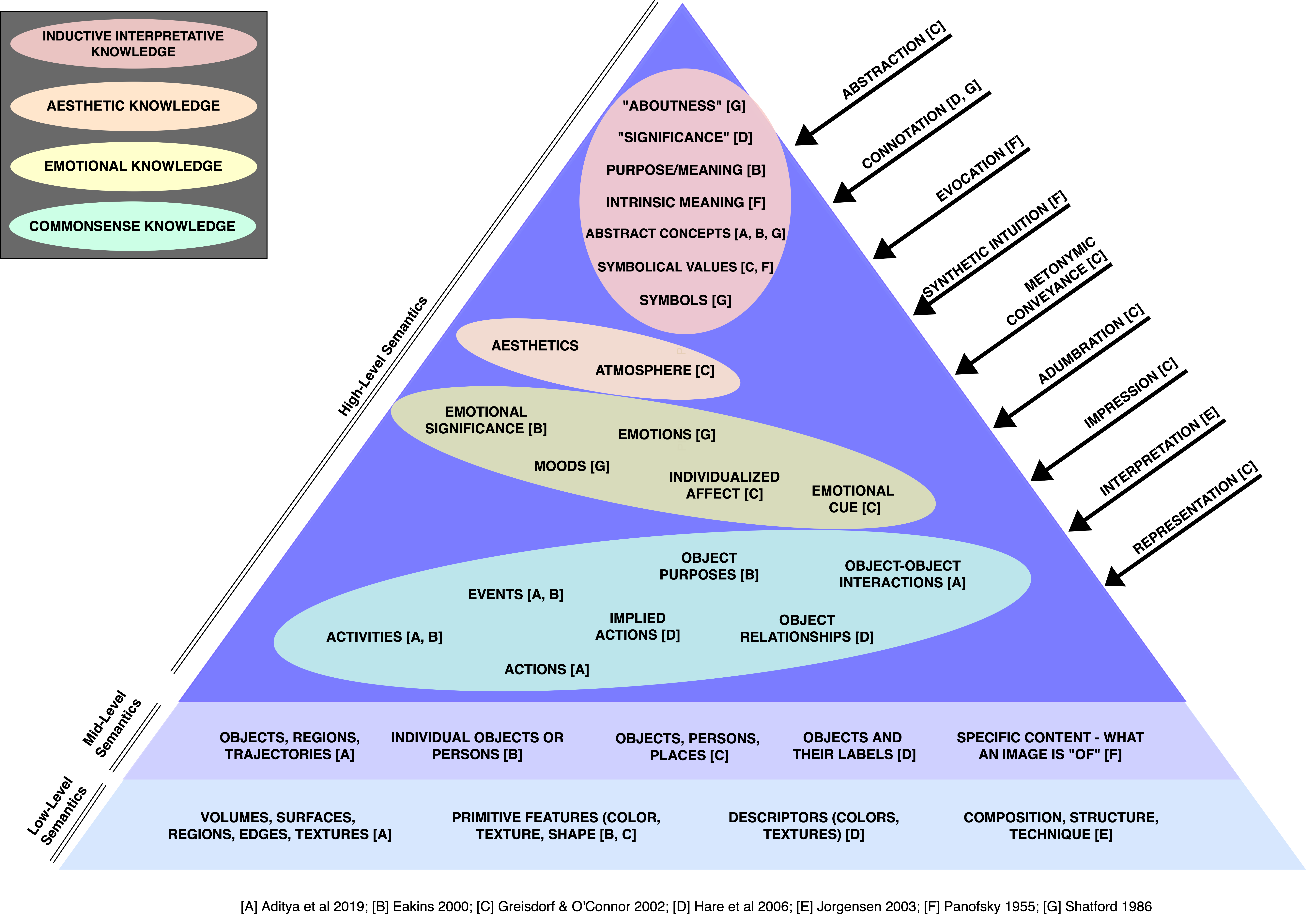}
        \caption[The tip of the iceberg: clusters of knowledge in high-level visual semantics.]{Tip of the iceberg: a deeper characterization of the top level of the visual semantic pyramid. Drawing from a multidisciplinary exploration of semantic entities associated with this upper semantic layer, we have identified four distinct clusters of knowledge.
        \label{fig:high-level}}
    \end{minipage}
\end{figure}

Part of the difficulty of solidifying a cross-disciplinary shared understanding of high-level semantics is that, in comparison to the other levels, high-level understanding by humans is increasingly cognitively complex. Complex cognitive processes, including abstraction, metonymic conveyance, adumbration, impression, prototypical displacement \cite{greisdorf2002modellingwhat}, connotation \cite{hare2006mind, shatford1986analyzingsubjecta}, evocation, and synthetic intuition \cite{panofsky1955meaning} are considered crucial tools for understanding visual semantics at this “high level” of abstraction. However, it is generally thought that it is practically hard to grasp them using typical automatic image understanding and indexing methods. As such, this highest level of abstraction in the interpretation of image meaning or content is seen as a “seemingly insurmountable obstacle” to the application of content-based image retrieval techniques \cite{enser1999visualimage}. 

In addition to cognitive complexity, subjectivity represents another challenging aspect when it comes to characterizing and automatically recognizing semantic units within this level. Shatford's widely cited insight encapsulates this notion succinctly: \textit{``...the delight and frustration of pictorial resources is that a picture can mean different things to different people"} \cite[p. 42]{shatford1986analyzingsubjecta}. Furthermore, a single picture can convey diverse meanings not only to various people but also to the same individual in different contexts or at different times of necessity. In line with this perspective, Greisdorf \cite{greisdorf2002modellingwhat} underscores the importance of interdisciplinary perspectives as a foundational approach for modeling the attributes of the human image cataloging process, because:

    \begin{quote}
    Those attributes tend to elude the indexing/cataloging process by exceeding the image indexing threshold due to individual viewer cognitive displacement of objects and object features that give rise to disjunctive prototypes that viewers may associate with the objects included as part of the image composition. These adumbrative, impressionistic and abstractionist concepts that relate viewer to image need to be captured with some type of retrieval mechanism in order to enhance retrieval effectiveness for system users. \cite[p. 11]{greisdorf2002modellingwhat}
    \end{quote}

To better comprehend and communicate about these abstract semantics, there is a need to precisely identify the semantic units that may belong to this layer and potentially characterize their interrelationships. Thus, we systematically reviewed the cited literature to provide a more detailed characterization of this apex of visual semantics (see Figure \ref{fig:high-level}). We categorized the types of elements mentioned as belonging to high-level visual semantics into four general groups:

\begin{itemize}
    \item \noindent \textbf{Commonsense semantics.} This cluster is closely aligned with mid-level semantics and is among the least subjective of the groups. It encompasses semantic elements such as explicit or implied actions (``running") \cite{aditya2019integrating, hare2006mind}, activities (``dance") \cite{aditya2019integrating, eakins2000retrieval}, events (``parade") \cite{aditya2019integrating, eakins2000retrieval}, relationships between objects or object-object interactions (``man holding cup") \cite{aditya2019integrating, hare2006mind}, and object purposes \cite{eakins2000retrieval}. These elements fall under the category of ``commonsense" knowledge because they often exhibit a high level of consensus among viewers and can be described within a logical framework.
    \item \noindent \textbf{Emotional semantics.} This cluster encompasses semantic information associated with emotions, encompassing moods, emotions \cite{shatford1986analyzingsubjecta}, emotional cues \cite{greisdorf2002modellingwhat}, emotional significance \cite{eakins2000retrieval}, and individualized affects \cite{greisdorf2002modellingwhat}.
    \item \noindent \textbf{Aesthetic semantics.} This smaller cluster focuses on global aesthetic attributes that pertain to the overall judgment of an image as a unified entity. Semantic units within this cluster include atmosphere \cite{greisdorf2002modellingwhat}.
    \item \noindent \textbf{Inductive interpretative semantics.} Positioned as the uppermost cluster, this group contains some of the most complex, subjective, and culturally encoded semantic elements within ``high-level semantics." It encompasses semantic units like an image's ``aboutness" \cite{shatford1986analyzingsubjecta}, significance \cite{hare2006mind}, purpose, and meaning \cite{eakins2000retrieval}, including intrinsic meaning \cite{panofsky1955meaning}. Crucially, this cluster includes symbols \cite{shatford1986analyzingsubjecta}, symbolical values \cite{greisdorf2002modellingwhat, panofsky1955meaning}, and abstract concepts \cite{aditya2019integrating, eakins2000retrieval, shatford1986analyzingsubjecta} as part of the top tier of visual semantics.
\end{itemize}

These clusters represent our initial effort to provide a preliminary characterization of the high-level layer of visual semantics. Although they are presented as distinct categories, there may be instances where semantic elements overlap between clusters, such as the intersection of mood and aesthetics, atmosphere and ACs, or emotion and ACs. Ongoing work may lead to further refinements and revisions of this diagram.

\subsection{Abstract Concepts and Visual Data}

In the cognitive science field, the term abstract concepts (ACs) \cite{borghi2018varietiesabstract, harpaintner2018semanticcontent, villani2019varietiesabstract, yee2019abstractionconcepts}, e.g. \textit{violence, freedom} or, \textit{danger}, refer to complex situations which do not possess a single and perceptually bounded object as referent, and which have more variable content both within and across individuals than concrete concepts \cite{barsalou2003abstractionperceptual, borghi2014wordssocial}. The mechanisms that underlie the formation and use of ACs are the object of study of the ``Words As social Tools" (WAT) cognitive theory \cite{borghi2014wordssocial, borghi2019wordssocial}, which sees words as tools to perform actions modifying the state of our social environment. In this sense, ACs are seen as tools that change the state of humans' inner processes, helping us formulate predictions and facilitating perception, categorization, and thought \cite{borghi2014wordssocial}.

As such, ACs are valuable for tasks like automatic indexing, retrieval, and managing visual data. Enser recognized over 20 years ago that concept-based image retrieval methods would continue to be vital for archival image collections \cite[p. 206]{enser1999visualimage}:

    \begin{quote}
    At the highest level of abstraction in the interpretation of image meaning or content, i.e. that which corresponds with Panofsky’s iconological level, the human reasoning based on tacit or world knowledge which underpins image indexing and retrieval operations poses a seemingly insurmountable obstacle to the application of CBIR techniques. At this level, we humans are able to ‘see’ within the primitive attributes of two dimensional imagery the portrayal of \textit{love, power, benevolence, hardship, discrimination, triumph, persecution and a host of other aspects of the human condition}. We are enabled, through the visual medium, to exercise skills in semiological analysis – the shared connotation of the icon, metonym and metaphor, the understanding and appreciation of two conceptually related but antithetical images
    \end{quote}

Importantly, the automatic association of ACs to images could lead to breakthroughs in a wide range of applications, including Web image search, online picture-sharing communities, and scientific and academic endeavors. It would benefit multiple applications by improving multimedia querying of digital libraries, as automatically generated ACs could be used for Boolean search, as traditional metadata are. Furthermore, enabling machines to recognize the potential of images to communicate ACs could be useful to CH institutions, to enhance their visual collections’ documentation, descriptions, and mediation--for example, using detected ACs to inform the design of multimodal interactive environments. Additionally, it could benefit public institutions building narratives about their visual objects to engage people from different backgrounds or with different abilities, companies in the creative sector exploiting their already existing catalogues, and companies building products or services related to specific abstract concepts.

Despite the considerable potential of ACs as descriptors of `aspects of the human condition' for visual indexing, the CV community has only recently started to tackle subjective and abstract content analysis \cite{hussain2017automaticunderstanding}. At first glance, challenges associated with high-level visual semantics become evident, including subjective perception influenced by personal and situational factors \cite{zhao2018affectiveimage}, a lack of shared methods and communication among researchers, class imbalances in popular datasets, and increased variability in query results involving abstract words \cite{abgaz2021methodologysemantic, kiela2014learningimage, lazaridou2015combininglanguage}. Furthermore, there is no explicit task definition or specific purpose datasets for AC image classification or AC detection from images. This lack of clarity necessitates an additional aspect of this survey: to explore past CV research that may have addressed the task implicitly, using different terminology.

\section{Survey Methodology}
\label{section:survey_method}

The primary objective of this survey is to identify CV methodologies, tools, and architectures designed for associating high-level semantic units, including ACs, with still images. In this section, we will outline our selection criteria for choosing papers within the CV field that address high-level abstract visual understanding tasks. This includes the criteria applied to select relevant publication venues and the keywords used for mining these repositories and venues. We will also elucidate the process for sub-selecting works that underwent additional detailed analysis for insights into AC image classification, and the criteria employed in the categorization of such works that implicitly or explicitly relate to the task.\\[5pt]

\noindent  \textbf{Venue Selection Criteria}
Our venue selection was guided by two key factors: relevance and impact. We initially narrowed our search to venues focused on CV, aligning with our research area's primary focus. Additionally, we assessed venue impact using established criteria to ensure high-quality and significant selections, thus confirming their suitability for our research. For journals, we initially exclusively considered those with a Q1 rating in the field of Computer Vision and Pattern Recognition, as determined by Scimago\footnote{\url{https://www.scimagojr.com/}. Access date: Janurary 2021.}. This criterion resulted in the identification of 19 journals and one book series. Notably, we excluded journals related to the medical field,\footnote{Journal of the Optical Society of America A: Optics and Image Science and Vision, International Journal of Computer Assisted Radiology and Surgery, Medical Image Analysis, and Computerized Medical Imaging and Graphics} leaving us with 15 selected journal venues. For conferences, we exclusively considered those with an A* or A rating according to CORE\footnote{\url{http://portal.core.edu.au/conf-ranks/}. Access date: January 2021.}. This  selection process yielded four conferences that met our impact criteria. The complete list of the 19 selected venues can be found in the Github repository.\\[5pt]

\begin{table}[h]
    \centering
    \small
    \resizebox{\textwidth}{!}{%
    \begin{tabular}{p{5cm}|p{9cm}}
    \hline
    \textbf{Category} & \textbf{Keywords} \\
    \hline
    \textbf{``Abstract" Adjectives} (synonyms, hyponyms of, or similar to ``abstract") & \textit{abstract, intangible, non-concrete, non-physical, symbolic, latent, evoked, implied, subjective, social, cultural, moral, political, economic, affective} \\
    \hline
    \textbf{``Concept" Nouns} (synonyms, hyponyms of, or similar to ``concept") & \textit{concept, object, idea, class, value, ideology, emotion, affect, sentiment, signal, attribute} \\
    \hline
    \textbf{CV Task Nouns} (synonyms, hyponyms of, or similar to “detection") & \textit{detection, classification, recognition, identification} \\
    \hline
    \end{tabular}%
    }
    \caption[Keywords for query construction]{Keywords for query construction, categorized into adjectives similar or synonymous with `abstract' and `high-level', nouns denoting similar ontological objects to `concepts', and task nouns. This categorization enables the creation of queries to identify works related to high-level concept detection or similar tasks.}
    \label{tab:keywords}
\end{table}

\noindent \textbf{Keywords and Query Building}. High-level visual sensemaking within CV lacks standardized nomenclature. The preliminary investigation into high-level semantic units in Section \ref{section:survey_tip} revealed the interchangeable use of various terms to describe abstract notions across multiple disciplines. Subsequently, we compiled a list of keywords, considering the diverse terminology used to denote these concepts and high-level abstract understanding across different fields. Our search on major academic databases for publications related to AC image classification from a CV perspective informed the final list of keywords. These keywords are categorized into adjectives, nouns, and nouns describing related tasks like detection or identification. By combining terms from these categories, we aimed to construct queries capable of identifying works related to AC image classification or similar tasks (see Table \ref{tab:keywords}). \\[5pt]

\noindent \textbf{High Level Visual Understanding Search}. The list of keywords was employed to construct search queries for the selected venues, with the inclusion of the term ``image."  All searches across the chosen venues were conducted using the SCOPUS\footnote{\url{https://www.scopus.com}. Access date: May 2022.} citation database. An exception was made for the ECCV Conference and Workshop, as they were not accessible through this database. The results were manually reviewed to collect papers specifically addressing high-level visual understanding of still images, i.e., those dealing with some of the semantic elements identified in Section \ref{section:survey_tip}. Through this process, 52 papers were gathered, comprising 6 journal papers (2017--2021) and 44 conference papers (2008--2021). Subsequently, a pruning phase was conducted to exclude articles meeting any of the following criteria:

\begin{itemize}
    \item Articles not written in English; only articles in the English language were considered for this survey.
    \item Articles describing high-level visual semantics tasks applied exclusively to moving images (e.g., videos), such as \cite{dibeklioglu2012areyou} and \cite{dasilva2020recognitionaffective}. However, articles that applied methods to both video and still images were considered.
    \item Articles addressing image captioning in a general context, without a focus on high-level reasoning tasks (e.g., \cite{li2019entangledtransformer}).
    \item Articles describing methods that relied on additional data sources beyond raw pixel data, particularly those dealing with multimodal or cross-modal data (e.g., text and images from news articles together, etc.), as in \cite{corchs2019ensemblelearning}, \cite{thomas2020preservingsemantic}, and \cite{vadicamo2017crossmedialearning}.
    \item Articles centered on image generation (e.g., \cite{raguram2008computingiconic}).
    \item Articles dealing with high-level visual semantics tasks, such as cultural event recognition, but produced within the context of the Chalearn LAP challenge (e.g., \cite{liu2015exploitingfeature}, \cite{rothe2015dldrdeep}, \cite{wang2015betterexploiting}, \cite{wei2015deepspatial}), as these have been previously surveyed \cite{escalera2015chalearnlooking}.
\end{itemize}

\noindent \textbf{General Survey Criteria:} This resulted in the selection of a total of 38 publications, comprising 4 journal papers and 34 conference papers. In our survey of these 38 publications, we aim to identify several key criteria, including:

\begin{itemize}
    \item \textbf{Image Type:} We examine whether the publications focus on natural photographs, art images, or other image types.
    
    \item \textbf{Related Semantic Elements:} We identify the high-level semantic units of interest, including examples, in each of the publications.

    \item \textbf{Datasets:} We analyze the datasets used and whether the authors created them for their research.
    
    \item \textbf{Computer Vision Tasks:} We categorize the explicit CV-related tasks addressed in each publication.
    
    \item \textbf{CV Task Clusters:} We classify each publication into relevant clusters of CV tasks.
\end{itemize}

\noindent \textbf{Subselection for In-Depth Surveying:} Upon reviewing the initial set of 38 publications, we observed a wide spectrum of high-level CV tasks with broad coverage of semantic units. To align with our core objective of studying ACs more closely aligned with values and ideologies, we then did an additional round of surveying for works addressing semantic units referred to as ``symbols" \cite{hussain2017automaticunderstanding, ye2018advisesymbolismb, ye2019interpretingrhetoric, kalanat2022symbolic}, ``intent" \cite{jia2021intentonomydataset}, and ``abstract topics" \cite{thomas2019predictingpolitics, thomas2021predictingvisual}. We did not consider works dealing with emotions, given the existence of a substantial number of surveys focused on them (e.g., \cite{ortis2020surveyvisual}). By narrowing our selection, we ensured methodological alignment with our research objectives. With a relatively limited number of qualifying publications, we extended our exploration using a bottom-up approach. By meticulously reviewing the bibliographic references of previously surveyed works, we identified additional publications, including interconnected or extended works. Notably, pairs like \cite{ye2019interpretingrhetoric} extending \cite{ye2018advisesymbolismb} and \cite{thomas2021predictingvisual} extending \cite{thomas2019predictingpolitics} were considered together in our in-depth survey. \\[5pt]

\noindent \textbf{Specific In-Depth Criteria.} For the 8 selected works, we conducted a more detailed analysis, classifying them based on various dimensions. In addition to the general dimensions, we introduced other criteria for our comprehensive survey:

\begin{itemize}
    \item \textbf{Explicit vs. Analogous/Related Task}: Explicit task addressed versus analogous AC image classification task.
    
    \item \textbf{Model Architecture}: Backbone architecture for the analogous task (e.g., CNN with perturbation, transformer with KG and GCN).
    
    \item \textbf{Reproducibility}: Availability of software and data.
    
    \item \textbf{Performance Metric}: Reported macro F1 score for the analogous task, if available.
    
    \item \textbf{System Hybridity}: Classification of the system as statistical, symbolic, or hybrid based on \cite{vanbekkum2021modulardesign}'s taxonomy.
\end{itemize}

\section{Automatic High-Level Visual CV Tasks}
\label{section:survey_ahlvu}


Our survey aimed to evaluate the contemporary advancements in Computer Vision (CV) for automatically detecting high-level semantic units in static images. Following the outlined methodology in Section \ref{section:survey_method}, we extensively searched top-rated publications in Computer Vision and Pattern Recognition. This endeavor yielded 38 notable publications, comprising 4 journal papers (2017-2021) and 34 conference articles (2008-2021). Table \ref{tab:cvtasks} presents an overview of our examination of selected works, aiming to uncover significant trends in the application of CV technologies for high-level visual semantic understanding. We explored various facets of these works, including their targeted image domains (e.g., natural photographs, art images), dataset utilization (and curation), and specific CV tasks addressed. Our primary focus was to identify the high-level semantic units addressed by each paper concerning static images, with illustrative examples provided where applicable. This thorough analysis provided valuable insights into the exploration of Abstract Concepts (ACs) within the existing literature.

Based on the table, distinct trends emerge, with a significant portion of this research gravitating toward specific CV tasks. These tasks include situation recognition \cite{Yatskar_2016_CVPR, suhail2019mixturekernelgraph, pratt2020grounded, li2017situationrecognition}, social relation recognition \cite{zhang2012reviewautomatic, wang2008surveyemotional, sun2017domainbased, li2020graphbasedsocial, li2017dualglancemodel, goel2019endtoendnetwork}, event recognition \cite{yao2019attentionawarepolarity, yuanjunxiong2015recognizecomplex, bossard2013eventrecognition}, visual persuasion and intent analysis \cite{joo2014visualpersuasion, jia2021intentonomydataset, huang2016inferringvisual, guo2021detectingpersuasive}, automatic advertisement understanding \cite{ye2019interpretingrhetoric, ye2018advisesymbolismb, hussain2017automaticunderstanding}, visual sentiment analysis \cite{yao2019attentionawarepolarity, toisoul2021estimationcontinuous, achlioptas2021artemisaffective}, aesthetic evaluations \cite{workman2017understandingmapping, gray2010predictingfacial, datta2006studyingaesthetics}, and analysis of social and personality traits \cite{segalin2017socialprofiling, masiprodo2012rolefacial, joo2015automatedfacial}, including occupation \cite{shao2013whatyou}, fashion \cite{kiapour2014hipsterwars, hsiao2017learninglatent}, and group-level analyses \cite{ghosh2019rolegroup, gallagher2009understandingimages}. Other CV tasks include abstract reasoning \cite{stabinger2017evaluation}, affordance reasoning \cite{chuang2018learningact}, political bias detection \cite{thomas2021predictingvisual}, visual humor detection \cite{chandrasekaran2016weare}, and environmental variables prediction \cite{khosla2014lookingvisible}. These task trends seem to align with the identified semantic units in Section \ref{section:survey_tip}, encompassing emotions, events, aesthetics, atmosphere, object purposes, object-object interactions, symbols, and more, reflecting the diverse landscape of high-level visual semantic exploration.

\begin{landscape}
    \tiny
    \begin{longtable}{p{0.02\linewidth}|p{0.02\linewidth}|p{0.12\linewidth}|p{0.12\linewidth}|p{0.025\linewidth}|p{0.16\linewidth}|p{0.15\linewidth}|p{0.1\linewidth}|p{0.12\linewidth}}
        \toprule
        \textbf{Year} &
        \textbf{Work} &
        \textbf{Image Type} &
        \textbf{Dataset} &
        \textbf{Own Data} &
        \textbf{Task(s)} &
        \textbf{Related Semantic Unit} &
        \textbf{Semantic Unit Examples} &
        \textbf{CV Cluster(s)} \\* \midrule
        \endhead
        2006 &
        \cite{datta2006studyingaesthetics} &
        natural photograph &
        Untitled &
        Y &
        aesthetic quality inference &
        aesthetics &
        \textit{aesthetic value} &
        aesthetic analysis \\* \midrule
        2009 &
        \cite{gallagher2009understandingimages} &
        natural photographs of groups of people &
        The Images of Groups Dataset &
        Y &
        demographic information detection; event recognition &
        events &
        \textit{dining, age, gender} &
        social signal processing; situational analysis \\* \midrule
        2010 &
        \cite{wang2010seeingpeople} &
        natural photographs of groups of people &
        The Images of Groups Dataset &
        N &
        social relationship recognition &
        object-object interaction &
        \textit{siblings, mother-child, husband-wife} &
        social signal processing; situational analysis \\* \midrule
        2010 &
        \cite{gray2010predictingfacial} &
        natural photographs of frontal female faces &
        Untitled &
        Y &
        female facial beauty prediction &
        aesthetics; abstract concept &
        \textit{female beauty} &
        aesthetic analysis; social signal processing \\* \midrule
        2012 &
        \cite{masiprodo2012rolefacial} &
        natural photographs of faces &
        Untitled &
        Y &
        social dimension trait detection &
        individualized affect; object purposes &
        \textit{dominance, aggressiveness, threatening} &
        social signal processing \\* \midrule
        2013 &
        \cite{shao2013whatyou} &
        natural photograph &
        Occupation Database &
        Y &
        occupation prediction &
        object purposes &
        \textit{waiter, customer, clergy} &
        social signal processing \\* \midrule
        2013 &
        \cite{bossard2013eventrecognition} &
        natural photograph &
        PEC Data Set &
        Y &
        event recognition &
        social events &
        \textit{roadtrip, halloween, christmas} &
        situational analysis \\* \midrule
        2014 &
        \cite{joo2014visualpersuasion} &
        natural photographs of politicians &
        Visual Persuasion Dataset &
        Y &
        communicative intent recognition; visual persuasion understanding &
        object purposes; purpose/meaning &
        \textit{trustworthy, powerful, energetic} &
        visual rhetorical analysis \\* \midrule
        2014 &
        \cite{khosla2014lookingvisible} &
        natural street view images &
        Untitled &
        Y &
        environment navigation &
        situational analysis; atmosphere &
        \textit{proximity to a McDonald's, crime rate} &
        situational analysis \\* \midrule
        2014 &
        \cite{kiapour2014hipsterwars} &
        natural photographs of full body outfits &
        Untitled &
        Y &
        fashion style classification &
        aesthetics &
        \textit{hipster, bohemian, preppy} &
        social signal processing \\* \midrule
        2015 &
        \cite{joo2015automatedfacial} &
        natural photographs of politicians &
        Visual Persuasion Dataset &
        Y &
        political affiliation detection; personality trait detection &
        individualized affect; object purposes &
        \textit{intelligence, competence, democrat affiliation} &
        social signal processing; visual rhetorical analysis \\* \midrule
        2015 &
        \cite{yuanjunxiong2015recognizecomplex} &
        natural photographs of events &
        WIDER &
        Y &
        event recognition &
        event &
        \textit{parade, press conference, meeting} &
        event recognition \\* \midrule
        2015 &
        \cite{zhang2015learningsocial} &
        natural photographs with visible faces &
        Social Relation Dataset &
        Y &
        social relationship recognition &
        object-object interaction &
        \textit{friendliness, warmth, attachment} &
        social signal processing; situational analysis \\* \midrule
        2016 &
        \cite{Yatskar_2016_CVPR} &
        natural photograph &
        Situnet &
        Y &
        situation recognition &
        actions; implied actions; object purposes; object relationships &
        \textit{obstacle, audience, source} &
        situation recognition \\* \midrule
        2016 &
        \cite{chandrasekaran2016weare} &
        clipart images &
        AVH (Abstract Visual Humor) Dataset &
        Y &
        visual humor prediction &
        atmosphere; meaning &
        \textit{humor/funninness} &
        visual rhetorical analysis \\* \midrule
        2016 &
        \cite{huang2016inferringvisual} &
        natural photographs of politicians &
        Visual Persuasion Dataset &
        N &
        communicative intent recognition; visual persuasion understanding &
        object purposes; purpose/meaning &
        \textit{trustworthy, powerful, energetic} &
        visual rhetorical analysis \\* \midrule
        2017 &
        \cite{segalin2017socialprofiling} &
        natural photograph &
        PsychoFlickr corpus &
        N &
        personality trait-based image classification &
        individualized affect; significance &
        \textit{neuroticism, extraversion, openness} &
        social signal processing \\* \midrule
        2017 &
        \cite{stabinger2017evaluation} &
        artificial chessboard images &
        Untitled &
        Y &
        symmetry task, identity task &
        object-object interaction &
        \textit{identity, symmetry} &
        situational analysis \\* \midrule
        2017 &
        \cite{hussain2017automaticunderstanding} &
        image advertisements &
        Ads Dataset &
        Y &
        automatic advertisement understanding; symbolism prediction &
        symbols; symbolical values &
        \textit{danger, beauty, death} &
        visual rhetorical analysis \\* \midrule
        2017 &
        \cite{hsiao2017learninglatent} &
        natural photographs (fashion images) &
        Style Embedding Dataset &
        Y &
        style embedding learning &
        aesthetics &
        \textit{hipster, bohemian, preppy} &
        social signal processing; aesthetic analysis \\* \midrule
        2017 &
        \cite{li2017dualglancemodel} &
        natural photograph &
        PISC &
        Y &
        social relationship recognition &
        object-object interaction; atmosphere &
        \textit{friends, couple, professional relationship} &
        social signal processing \\* \midrule
        2017 &
        \cite{sun2017domainbased} &
        natural photograph &
        PIPA-relation &
        Y &
        social relationship recognition &
        object-object interaction; atmosphere &
        \textit{friends, colleagues, lovers} &
        social signal processing \\* \midrule
        2017 &
        \cite{workman2017understandingmapping} &
        ground-level and overhead natural photographs &
        ScenicOrNot (SoN) dataset &
        Y &
        scenicness prediction &
        atmosphere; aesthetics &
        \textit{scenicness of natural landscapes} &
        aesthetic analysis \\* \midrule
        2017 &
        \cite{li2017situationrecognition} &
        natural photograph &
        imSitu &
        N &
        situation recognition &
        actions; implied actions; object purposes; object relationships &
        \textit{obstacle, victim, destination} &
        situation recognition \\* \midrule
        2018 &
        \cite{chuang2018learningact} &
        natural photographs of scenes &
        ADE-Affordance &
        Y &
        affordance reasoning in scenes &
        object purposes &
        \textit{socially awkward affordance, socially forbidden affordance} &
        situation recognition \\* \midrule
        2018 &
        \cite{ye2018advisesymbolismb} &
        image advertisements &
        Ads Dataset &
        Y &
        automatic advertisement understanding &
        symbols; symbolical values &
        \textit{freedom, happiness, adventure} &
        visual rhetorical analysis; situational analysis \\* \midrule
        2019 &
        \cite{yao2019attentionawarepolarity} &
        natural photographs, art images and art paintings &
        Flickr and Instagram (FI); IAPSa; ArtPhoto; Abstract &
        N &
        affective image retrieval &
        emotion &
        \textit{contentment, awe, excitement, disgust} &
        visual sentiment analysis \\* \midrule
        2019 &
        \cite{suhail2019mixturekernelgraph} &
        natural photograph &
        imSitu &
        N &
        situation recognition, role assignment &
        actions; implied actions; object purposes; object relationships &
        \textit{audience, obstacle, source} &
        situation recognition \\* \midrule
        2019 &
        \cite{ghosh2019rolegroup} &
        natural photographs of groups of people &
        GAF-personage database &
        Y &
        most influential person identification &
        affect; object-object interactions &
        \textit{leader, influential person} &
        social signal processing \\* \midrule
        2019 &
        \cite{ye2019interpretingrhetoric} &
        image advertisements &
        Ads Dataset &
        Y &
        automatic advertisement understanding &
        symbols; emotions; abstract concepts &
        \textit{danger, adventure, death} &
        visual rhetorical analysis \\* \midrule
        2019 &
        \cite{goel2019endtoendnetwork} &
        natural photograph &
        PIPA-relation; PISC &
        N &
        social relationship recognition &
        object-object interaction; atmosphere &
        \textit{friends, couple, professional relationship} &
        social signal processing \\* \midrule
        2020 &
        \cite{li2020graphbasedsocial} &
        natural photograph &
        PIPA-relation; PISC &
        N &
        social relationship recognition &
        object-object interaction &
        \textit{colleagues, lovers, audience} &
        social signal processing \\* \midrule
        2020 &
        \cite{pratt2020grounded} &
        natural photograph &
        SWiG &
        Y &
        grounded situation recognition &
        actions; implied actions; object purposes; object relationships &
        \textit{obstacle, victim, destination} &
        situation recognition \\* \midrule
        2021 &
        \cite{achlioptas2021artemisaffective} &
        artwork images &
        ArtEmis &
        Y &
        affective image captioning; dominant emotion prediction &
        feelings; abstract concepts &
        \textit{awe, freedom, love} &
        visual sentiment analysis \\* \midrule
        2021 &
        \cite{guo2021detectingpersuasive} &
        image advertisements &
        Ads Dataset &
        N &
        persuasive atypicality detection &
        object purposes; object-object interactions &
        \textit{atypicality} &
        visual rhetorical analysis; situational analysis \\* \midrule        
        2021 &
        \cite{jia2021intentonomydataset} &
        natural photographs of everyday scenes &
        Intentonomy &
        Y &
        intent recognition &
        object purposes; purpose/meaning &
        \textit{harmony, mastery, perseverance} &
        visual rhetorical analysis \\* \midrule
        2021 &
        \cite{thomas2021predictingvisual} &
        images from news sources &
        Politics &
        Y &
        political bias prediction &
        symbolical values; intrinsic meaning &
        \textit{diversity, tradition, homelessness} &
        visual rhetorical analysis \\* \midrule
        2021 &
        \cite{toisoul2021estimationcontinuous} &
        natural photographs of faces &
        AffectNet, SEWA, AFEW-VA &
        N &
        categorical emotion recognition; valence estimation; arousal estimation &
        emotion &
        \textit{anger, calming, depressed} &
        visual sentiment analysis; social signal processing 
        \\* \bottomrule
        
        \caption[Survey of works in top CV venues that automatically detect high-level semantics from still images.]{Overview of surveyed research articles from top CV venues, focusing on high\-level semantic analysis tasks in still images. The table includes information on the year of publication, the work's title, image type, dataset used, whether the work used its own data, the specific tasks addressed, related semantic units, examples of semantic units, and the CV clusters associated with these tasks. The table offers a comprehensive look at various high\-level semantic analysis tasks, ranging from aesthetic quality inference to political bias prediction, and the semantic units and clusters involved in these tasks.}

    \end{longtable}
    \label{tab:cvtasks}

\end{landscape}

\subsection{Clustering High-Level CV Tasks}

\begin{figure}[!h]
    \centering
    \includegraphics[width=\linewidth]{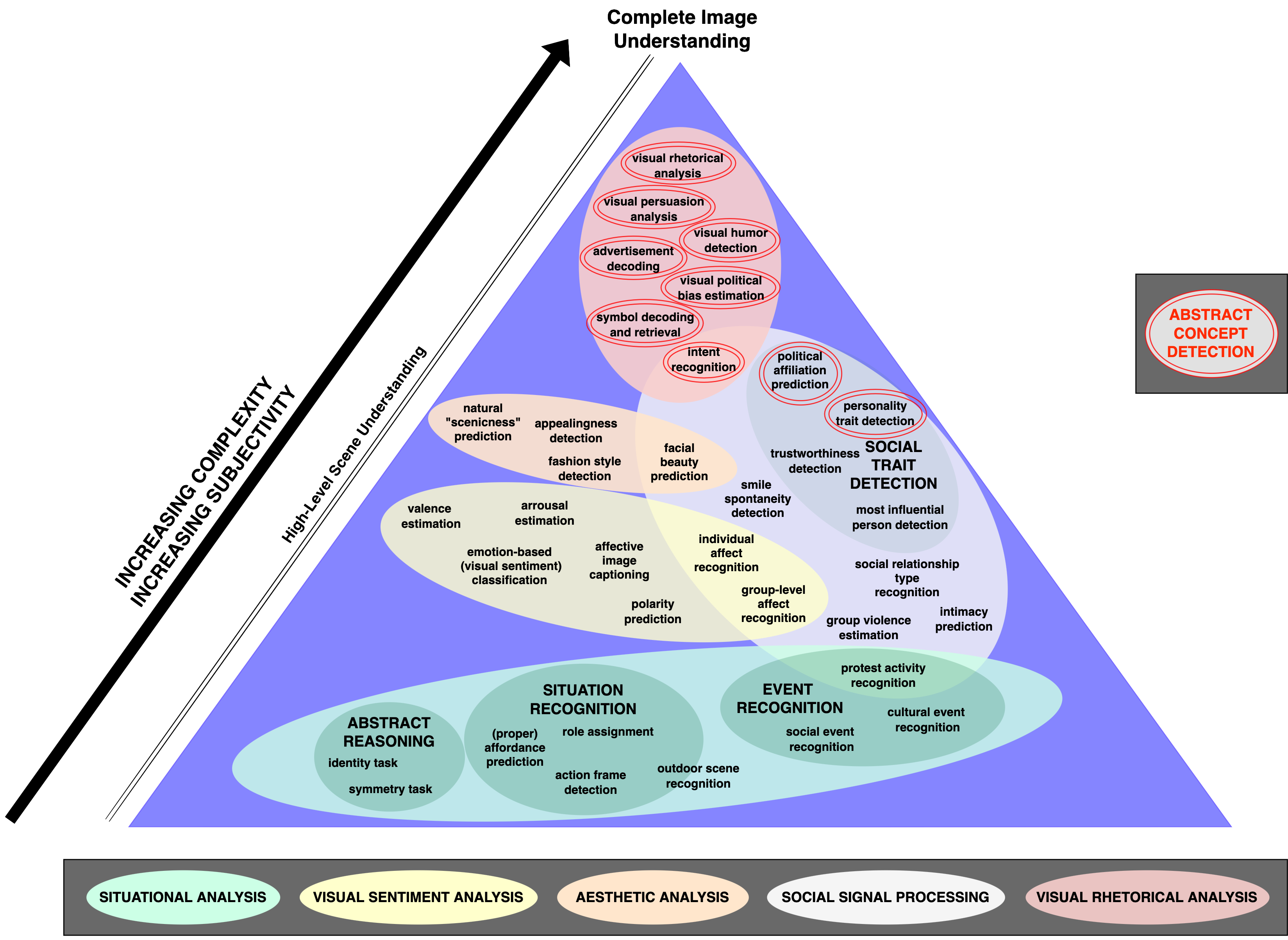}
    \caption[Clusters of CV tasks dealing with high-level visual semantics.]{Computer Vision tasks that deal with ``high level semantics" or ``high level visual understanding", which have been mapped also to the previous multidisciplinary characterization of high level semantics. Circled in red are the tasks that were found to implicitly or explicitly deal with AC detection.}
    \label{fig:high_cv}
\end{figure}

Based on these findings, we formulated a set of clusters for CV tasks associated with high-level semantic units, and we assigned each publication to one or more of these clusters (as shown in the last column of Table \ref{tab:cvtasks}). Subsequently, we leveraged these outcomes to construct a cluster-based diagram illustrating CV tasks linked to high-level visual reasoning (see Fig. \ref{fig:high_cv}. This diagram partly mirrors the structure of our high-level visual semantic unit diagram (Fig. \ref{fig:high-level}). In this way, the diagram serves as an alignment of the extensive body of work in CV pertaining to high-level visual reasoning and visual semantics with the same conceptual and terminological framework expounded upon in Section \ref{section:survey_tip}. We identified five main clusters of related CV tasks that deal with high-level semantic units:

\begin{itemize}
    \item \noindent \textbf{Situational analysis.} This cluster aligns with the commonsense semantics category, focusing on actions, activities, roles, and object purposes, along with the logical or common interactions among these elements. Within situational analysis, we can discern three primary sub-clusters of CV tasks. Firstly, abstract reasoning tackles global semantic tasks rooted in logic, such as assessing the symmetry of a chessboard or identifying its identity in various images, as exemplified in \cite{stabinger2017evaluation}. The second sub-cluster, situation recognition, as initially articulated by \cite{Yatskar_2016_CVPR}, revolves around summarizing the content of an image comprehensively. This entails identifying the primary activity, involved actors, objects, substances, locations, and most notably, the roles these participants assume within the activity. This structured prediction task goes beyond merely predicting the most prominent action, aiming to forecast the verb and its frame, which consists of multiple role-noun pairs. Finally, a substantial body of work is dedicated to event recognition, with a specific focus on social and cultural events.

    \item \noindent \textbf{Visual sentiment analysis.} This cluster aligns with the previously identified emotional semantics category and is also referred to as Image Emotion Analysis (IEA). Its primary goal is to comprehend how images elicit emotional responses in individuals. Despite being relatively recent, this subdomain has witnessed substantial growth in recent years, resulting in an extensive body of research and several surveys \cite{ortis2020surveyvisual, zhao2021computationalemotion}. Most studies have concentrated on emotion detection, aiming to identify emotions like fear, sadness, excitement, and contentment within natural images (as surveyed by \cite{zhao2021computationalemotion}). Some research has delved into the analysis of group emotions in images, as outlined in \cite{veltmeijer2021automaticemotion}. In addition to natural images, this field has explored other visual media types. Notably, automatic emotion detection has been explored in the context of art images \cite{cetinic2019deeplearning, achlioptas2021artemisaffective} and memes \cite{sharma2020semeval2020task}.

    \item \noindent \textbf{Aesthetic analysis.} This cluster aligns with the previously identified aesthetic semantics category. Although relatively limited in volume, research within this cluster predominantly revolves around the detection or prediction of aesthetic value in images. These investigations encompass a range of image types, including natural scenes \cite{workman2017understandingmapping}, images of human faces \cite{gray2010predictingfacial}, and images as a whole \cite{datta2006studyingaesthetics}.
    
    \item \noindent \textbf{Social signal processing.} This distinct cluster does not directly align with any of the previously identified clusters of the high-level semantics pyramid. It encompasses research related to Social Signal Processing, a broad field dedicated to constructing computational models for sensing and understanding various human social signals, including emotions, attitudes, personalities, skills, roles, and other forms of human communication. Within this cluster, several endeavors utilize CV techniques to detect non-concrete social aspects within images. In the realm of natural images, efforts have delved into discerning persuasive intent in political photographs \cite{joo2014visualpersuasion}, identifying deception \cite{bruer2020identifyingliars, zanette2016automateddecoding}, categorizing social relationship types (e.g., kinship, friendship, romantic, and professional relationships) \cite{li2020visualsocial}, and gauging intimacy \cite{chu2015multitaskrecurrent}. Group or crowd images have garnered particular attention, with various methodologies developed to automatically recognize non-concrete group attributes, including social cohesion \cite{hung2010estimatingcohesion, bkantharaju2020multimodalanalysis}, leadership \cite{solera2017groupsleaders}, warmth, and dominance \cite{zhang2018facialexpression}, excitement \cite{varghese2018deeplearning}, and engagement in student groups \cite{vanneste2021computervision}. However, social signal processing tasks in the context of art images remain relatively underexplored, with some research focusing on automatically detecting the trustworthiness of depicted individuals in art images \cite{safra2020trackinghistorical}.

    \item \noindent \textbf{Visual rhetorical analysis.}  This cluster exhibits a strong correlation with inductive interpretative semantics, focusing on understanding the intrinsic and implicit meaning and the purpose of images, akin to the concept of inductive interpretative semantics. It encompasses research related to visual persuasion and rhetorical techniques. Notably, much of this work has been concentrated in the subdomain of automatic advertisement understanding. While the visual rhetoric of images has traditionally been explored in Media Studies, the computational study of this field has gained prominence in the last five years, with a significant contribution from the authors of \cite{hussain2017automaticunderstanding}. Within this cluster, one notable sub-cluster revolves around visual persuasion \cite{joo2014visualpersuasion}. This subfield assesses whether images of politicians present them in a positive or negative light by analyzing facial expressions, gestures, and image backgrounds, using features to discern the persuasive intent behind the visuals.
\end{itemize}

It is worth noting that categorizing certain works into distinct clusters can be challenging, as some research efforts overlap between these domains. For instance, \cite{ghosh2019rolegroup} explores the impact of group-level affect in identifying the most influential person within images of groups. This study straddles the subdomains of both visual sentiment analysis and social signal processing, highlighting the interconnected nature of these research areas.

\subsection{Discussion on High-Level CV Tasks}

\begin{table*}[!h]

    \footnotesize
  \begin{tabularx}{\textwidth}{p{0.25\textwidth}|p{0.44\textwidth}|p{0.45\textwidth}}
    \toprule
    Semantic Unit Type & Examples & References\\
    \midrule
    \textit{events} & dining, roadtrip, parade & \cite{gallagher2009understandingimages, bossard2013eventrecognition, yuanjunxiong2015recognizecomplex} \\
    \textit{relationship type} & siblings, colleagues, lovers  & \cite{wang2010seeingpeople, li2017dualglancemodel, sun2017domainbased, goel2019endtoendnetwork, li2020graphbasedsocial} \\
    \textit{relationship intimacy} & friendliness, warmth  & \cite{zhang2015learningsocial} \\
    \textit{social dimension} & leader, influential person  & \cite{ghosh2019rolegroup} \\
    \textit{personality trait} & dominance, competence, neuroticism  & \cite{masiprodo2012rolefacial, joo2015automatedfacial, huang2016inferringvisual, segalin2017socialprofiling} \\
    \textit{communicative intent} & trustworthy, powerful, mastery, harmony  & \cite{huang2016inferringvisual, jia2021intentonomydataset} \\
    \textit{occupation} & waiter, clergy  & \cite{shao2013whatyou} \\
    \textit{political affiliation} & democrat, republican  & \cite{joo2014visualpersuasion} \\
    \textit{political bias} & liberal, conservative  & \cite{thomas2021predictingvisual} \\
    \textit{aesthetics} & female beauty, aesthetic value & \cite{datta2006studyingaesthetics, gray2010predictingfacial} \\
    \textit{atmosphere} & crime rate, scenicness  & \cite{khosla2014lookingvisible, workman2017understandingmapping} \\
    \textit{fashion style} & hipster, bohemian  & \cite{kiapour2014hipsterwars, hsiao2017learninglatent} \\
    \textit{humor} & humor, funninness  & \cite{chandrasekaran2016weare} \\
    \textit{object purposes} & obstacle, audience, destination, affordance  & \cite{Yatskar_2016_CVPR, li2017situationrecognition, chuang2018learningact, suhail2019mixturekernelgraph, pratt2020grounded} \\
    \textit{object interaction} & identity, symmetry  & \cite{stabinger2017evaluation} \\
    \textit{symbolical values} & danger, comfort, freedom  & \cite{hussain2017automaticunderstanding, ye2018advisesymbolismb, thomas2021predictingvisual, ye2019interpretingrhetoric} \\
    \textit{rhetorics} &  atypicality  & \cite{guo2021detectingpersuasive} \\
    \bottomrule
  \end{tabularx}
    \caption{Types of High-Level Semantics Extracted from Images }
  \label{tab:commands}
\end{table*}

\begin{table*}[!h]

    \footnotesize
  \begin{tabularx}{\textwidth}{p{0.18\textwidth}|p{0.52\textwidth}|p{0.23\textwidth}}
    \toprule
    Image Type & Datasets & References \\
    \midrule
    \textit{natural images} & ImageNet \cite{deng2009imagenet}, Occupation Database \cite{shao2013whatyou}, PEC Data Set \cite{bossard2013eventrecognition}, Situnet a.k.a imSitu \cite{Yatskar_2016_CVPR}, PyschoFlickr \cite{cristani2013unveiling}, PISC \cite{li2020visualsocial}, PIPA-relation \cite{sun2017domainbased}, ScenicOrNot \cite{workman2017understandingmapping}, ADE-Affordance \cite{chua2009nuswiderealworld}, SWiG \cite{pratt2020grounded}, Intentonomy \cite{jia2021intentonomydataset}, WIDER \cite{yuanjunxiong2015recognizecomplex} & \cite{datta2006studyingaesthetics, shao2013whatyou, bossard2013eventrecognition, khosla2014lookingvisible, Yatskar_2016_CVPR, segalin2017socialprofiling, li2020visualsocial, sun2017domainbased, workman2017understandingmapping, li2017situationrecognition, chuang2018learningact, suhail2019mixturekernelgraph, goel2019endtoendnetwork, li2020graphbasedsocial, pratt2020grounded, jia2021intentonomydataset, yuanjunxiong2015recognizecomplex} \\
    \textit{facial images} & AffectNet \cite{mollahosseini2017affectnet}, SEWA \cite{kossaifi2019sewa}, Social Relation Dataset \cite{zhang2015learningsocial} & \cite{joo2015automatedfacial, gray2010predictingfacial, masiprodo2012rolefacial, zhang2012reviewautomatic, toisoul2021estimationcontinuous} \\
    \textit{fashion images} & Style Embedding Dataset \cite{hsiao2017learninglatent} & \cite{kiapour2014hipsterwars, hsiao2017learninglatent} \\
    \textit{group images} & Images of Groups \cite{gallagher2009understandingimages}, GAF-personage \cite{ghosh2019rolegroup} & \cite{gallagher2009understandingimages, wang2010seeingpeople, ghosh2019rolegroup} \\
    \textit{political images} & Visual Persuasion \cite{joo2014visualpersuasion}, Politics \cite{thomas2021predictingvisual} & \cite{joo2015automatedfacial, joo2014visualpersuasion, huang2016inferringvisual, thomas2021predictingvisual} \\
    \textit{advertisements} & Ads Dataset \cite{hussain2017automaticunderstanding} & \cite{hussain2017automaticunderstanding, guo2016jointlyembedding, ye2018advisesymbolismb, ye2019interpretingrhetoric} \\
    \textit{artworks} & ArtEmis \cite{achlioptas2021artemisaffective} & \cite{achlioptas2021artemisaffective} \\
    \textit{clipart} & AVH Dataset \cite{chandrasekaran2016weare} & \cite{chandrasekaran2016weare} \\
    \bottomrule
  \end{tabularx}
    \caption{Types of Images Analyzed for High Level Semantics}
  \label{tab:commands}
\end{table*}

\subsubsection{Social and Sociocultural Emphasis in High-Level Visual Semantics}

The survey encompasses a wide array of semantic units (see Table \ref{tab:commands}), but social aspects, including emotions, relationships, and social events, constitute one of the most recurrent themes. This emphasis on human-centered concepts underscores the growing significance of sociocultural elements in high-level visual semantics research, and highlights a substantial and growing body of research in CV that focuses on social aspects like social relationship recognition, visual rhetorical analysis, and situational analysis. This signifies a notable shift towards understanding the intricate interplay between visual content and societal contexts, reflecting the increasing importance of sociocultural elements in this research domain.

\subsubsection{Diversifying Image Types} 
While the majority of surveyed works (approximately 75\%) predominantly revolve around natural photographs, shedding light on the importance of real-world imagery in high-level semantic research, a notable inclusion of publications dealing with cultural images stands out. These cultural images span artistic, historical, and advertisement domains, hinting at an evolving inter- or cross-disciplinary approach within these types of CV tasks. This evolving trend underscores the growing relevance of cultural images and signifies that CV is extending its reach beyond conventional photography, engaging with diverse and culturally significant visual contexts.

\subsubsection{Task-Specific Dataset Creation} 
Particularly noteworthy is the finding that an overwhelming majority, approximately 75\%, of studies exploring high-level semantics have embarked on the creation of bespoke datasets. This observation hints at the inherent complexities in attempting to generalize across a wide spectrum of cognitively intricate tasks, where the one-size-fits-all approach of general-purpose datasets may fall short. This trend reflects the intricate and task-specific nature of endeavors related to high-level semantic units, often requiring datasets customized to address the nuances of the research objectives.

\subsubsection{Research Output and Transformative Moments}

\begin{figure}[!h]

        \centering
        \includegraphics[width=.8\textwidth]{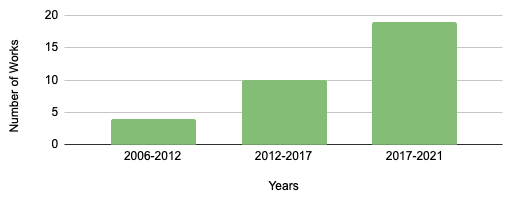}
        \caption[Inflection points for CV publications of high-level visual understanding.]{Two inflection points, (2012) and (2017), that seem to correlate with the increasing interest in CV tasks dealing with high-level visual semantics.}
    \label{fig:trend}
\end{figure}

The survey results demonstrate a pronounced rise in publications concerning high-level semantic units in top CV venues over a 15-year duration. Substantial increases in these types of publications in top CV venues seem to coincide with two pivotal moments for the CV field, 2012 and 2017 (see Figure \ref{fig:trend}). Specifically, our results suggest a potential correlation between higher interest in high-level tasks in CV research and the rise of DL around 2012, impulsed by the introduction of AlexNet \cite{krizhevsky2012imagenet}. Before this breakthrough, CV predominantly relied on traditional ML techniques and manual feature engineering, facing difficulties in high-level semantic tasks. However, the advent of deep neural networks revolutionized the field by enabling autonomous learning of hierarchical features from data, and Convolutional Neural Networks (CNNs) played a pivotal role in these developments, empowering the field to tackle intricate aspects of visual semantics. Our survey specifically underscores a substantial increase in publications related to high-level visual understanding post-2012 which coincides with the greater trend in the field of CV.  Building on this momentum, CV entered another transformative phase in 2017, marked by significant advancements in various subfields. This period expanded the scope of high-level semantic tasks, driven by innovations such as YOLO (You Only Look Once) and its improvements \cite{redmon2016you, redmon2017yolo9000}, which revolutionized real-time object detection, Mask R-CNN \cite{he2017mask}, which extended the popular Faster R-CNN framework to enable instance segmentation to identify pixel-level object boundaries, and progress in pose estimation \cite{kendall2017geometric}, among others. The strong increase in publications after 2017 suggests that researchers in CV recognized that they now possessed the tools and methodologies to delve deeper into complex abstract semantics within visual data, propelling the field to tackle previously challenging tasks.

\section{In-Depth Survey of ACs in CV}
\label{section:survey_ac_ic}

As explained in Section \ref{section:survey_method}, to align our research focus with the association of ACs to visual data, we initiated a systematic subselection process to identify the works most closely dealing with tasks analogous to AC image classification. We specifically sought works emphasizing ACs as socially shared meanings embodying values and ideologies, so we did a targeted approach to study works addressing semantic units termed ``symbols," ``intent," and ``abstract topics." By narrowing our selection, we identified 8 closely related works and classified them based on various dimensions. These dimensions included explicit versus analogous/related tasks, architectural models, reproducibility, macro F1 scores, and the hybridity of the systems. The results can be found in Tables \ref{tab:ac_ic_overall} and \ref{tab:ac_ic_arch}.

    \begin{table}[ht]
    \centering
    \scriptsize
        \begin{tabular}{p{0.035\linewidth}|p{0.07\linewidth}|p{0.34\linewidth}|p{0.04\linewidth}|p{0.1\linewidth}|p{0.04\linewidth}|p{0.04\linewidth}|p{0.12\linewidth}}
        \textbf{Year} & \textbf{Work} & \textbf{Title} & \textbf{Img Type} & \textbf{Dataset} & \textbf{Own Data} & \textbf{Data Size} & \textbf{AC Examples} \\
        \hline
        2016 & \cite{ahresabstractconcept} & \textit{Abstract Concept and Emotion Detection in Tagged Images with CNNs} & SM & NUS-WIDE & $\times$ & 14K & \textit{love, travel, beauty} \\
        2017 & \cite{hussain2017automaticunderstanding} & \textit{Automatic understanding of image and video advertisements} & Ad & Ads Dataset & $\checkmark$ & 14K & \textit{danger, fun, beauty} \\
        2019 & \cite{ye2018advisesymbolismb, ye2019interpretingrhetoric} & \textit{Interpreting the Rhetoric of Visual Advertisements} & Ad & Ads Dataset & $\checkmark$ & 14K & \textit{danger, fun, beauty}  \\
        2021 & \cite{thomas2019predictingpolitics, thomas2021predictingvisual} & \textit{Predicting Visual Political Bias Using Webly Supervised Data} & P & Politics & $\checkmark$ & 1M & \textit{abortion, justice, democrat} \\
        2021 & \cite{jia2021intentonomydataset} & \textit{Intentonomy: A dataset and study towards human intent understanding} & SM & Intentonomy & $\checkmark$ & 14K & \textit{harmony, power, beauty} \\
        2022 & \cite{kalanat2022symbolic} & \textit{Symbolic image detection using scene and knowledge graphs} & Ad & Ads Dataset & $\times$ & 8K & \textit{danger, fun, beauty} \\
        \end{tabular}
        \caption{Overview of CV studies and associated datasets for tasks closely resembling AC image classification. Ad: Advertisement, SM: Social Media, P: Political.}
        \label{tab:ac_ic_overall}

    \end{table}

    \begin{table}[!ht]
    \centering
    \scriptsize
    \begin{tabular}{p{0.035\linewidth}|p{0.07\linewidth}|p{0.13\linewidth}|p{0.12\linewidth}|p{0.25\linewidth}|p{0.04\linewidth}|p{0.06\linewidth}|p{0.09\linewidth}}
    \textbf{Year} & \textbf{Work} & \textbf{Explicit Task} & \textbf{Related Task} & \textbf{Model} & \textbf{Rep.} & \textbf{Macro F1} & \textbf{Hybridity} \\
    \hline
    2016 & \cite{ahresabstractconcept} & AC detection & AC detection & CNN with tag-specific binary classifiers & $\times$ & 0.18 & Statistical \\
    2017 & \cite{hussain2017automaticunderstanding} & \makecell[tl]{Automatic ads \\ understanding} & \makecell[tl]{Symbolism \\ detection} & CNN with region attention-based image classifier & $\checkmark$ & 0.16 & Statistical \\
    2019 & \cite{ye2018advisesymbolismb, ye2019interpretingrhetoric} & \makecell[tl]{Automatic ads \\ understanding} & \makecell[tl]{Symbolism \\ detection} & Undefined classifier and knowledge base & $\checkmark$ & -- & Hybrid \\
    2021 & \cite{thomas2019predictingpolitics, thomas2021predictingvisual} & \makecell[tl]{Visual political  \\ bias detection} & Image-word alignment & CNN multimodal feature learner, visual-only test time & $\checkmark$ & 0.25* & Statistical \\
    2021 & \cite{jia2021intentonomydataset} & \makecell[tl]{Intent \\ recognition} & \makecell[tl]{Intent \\ detection} & CNN with perturbation approach & $\checkmark$ & 0.23 & Hybrid \\
    2022 & \cite{kalanat2022symbolic} & \makecell[tl]{Symbolic image \\ classification} & \makecell[tl]{Symbolism \\ detection} & Transformer with KG, CGN, and attention & $\checkmark$ & 0.15 & Hybrid \\
    \end{tabular}
    \caption{In-depth examination of model architectures and performance in CV work analogous to AC image classification. AC: Abstract Concept. CNN: Convolutional Neural Network. GCN: Graph Convolutional Network.}
    \label{tab:ac_ic_arch}
    \end{table}

\paragraph{Overlap in Abstract Concept Examples} The AC examples employed across the surveyed works exhibit striking similarity. However, it is crucial to highlight that not all works share identical target classes due to variations in vocabulary and task definition. This divergence in target vocabulary hampers the ability to leverage the same datasets efficiently. There is evidently a demand for a shared dataset that compiles culturally rich images tagged with a common set of ACs. Such a resource could benefit future research by promoting data consistency and facilitating cross-work comparisons.

\paragraph{Diversity in Image Types} While the broader category of high-level visual understanding includes a prevalence of natural images, we observe a more concentrated focus on images with strong socio-cultural connotations or layers in works dealing with ACs. Notably, these images predominantly encompass social media content, advertisements, and political imagery. What sets these images apart is their cultural richness, as they often convey nuanced socio-cultural information that standard natural images may lack.

\paragraph{Emphasis on Advertisements} It is noteworthy that a significant portion of the works are centered around advertisements and employ the same dataset \cite{hussain2017automaticunderstanding, ye2018advisesymbolismb, yee2019abstractionconcepts, kalanat2022symbolic}. This suggests that the domain of advertisements is a good domain for the task and provides a rich source of data for the exploration of ACs in images.

\paragraph{Creation of Domain-Specific Datasets} Many of the surveyed works take the initiative to collect and present their own datasets. This underlines the critical importance of domain-specific data for the task of AC image classification.

\paragraph{Consistent Dataset Magnitude} Most works in this survey operate with datasets ranging from 8,000 to 14,000 images. It is important to note that the dataset size remains relatively consistent across these works. However, there is an exception in the case of \cite{thomas2019predictingpolitics}, which generates an extensive 1 million-image dataset for their primary research focus, but in a weakly supervised way. Critically, the dataset size for the subtask of image-word alignment is not explicitly reported.

\paragraph{Comparable Related Tasks} It is evident that many of the related tasks are comparable. They often share similar target concepts, such as \textit{beauty.} This observation indicates that there is a degree of research overlap between these tasks. This connection, although present, had not been explicitly highlighted before.

\paragraph{Dominance of CNN Architectures} Out of the six approaches analyzed, four employ Convolutional Neural Network (CNN) architectures as the backbone of their model structures. This prevalence suggests that deep learning paradigms, particularly CNNs, constitute a cornerstone of state-of-the-art AC image classification tasks.

\paragraph{Significance of F1 Score and Reporting Challenges} Within this body of work, it is evident that the F1 score holds greater importance as a model evaluation metric than accuracy. This underscores the essential role of the F1 score as the primary performance measure in the realm of AC image classification. However, a noteworthy challenge emerges concerning the reporting of these metrics. In specific instances, the analogous task for some works exists as a subtask within the broader context of the work's primary objective. For example, symbolism prediction constitutes a subtask within the larger framework of understanding visual advertisements. As a result, performance scores related to the analogous tasks may not be explicitly reported for all works, as observed in \cite{ye2018advisesymbolismb, ye2019interpretingrhetoric}. Notably, this reporting gap posed a particular challenge when evaluating the highest F1 score work, namely \cite{thomas2021predictingvisual}, where we had to compute the score ourselves.\footnote{To achieve this, we calculated the average F1 score based on the reported F1 scores for the analogous image-word alignment tasks. We excluded scores related to the names of politicians and media outlets due to their concreteness, specifically: 'cnn,' 'trump,' 'clinton,' 'donald,' 'paul,' 'fox,' 'clinton,' 'hillary,' 'obama,' and 'republican.' The resulting F1 score was calculated based on the average scores for the following words: 'administration,' 'political,' 'conference,' 'meeting,' 'prime,' 'committee,' 'host,' 'minister,' 'foreign,' 'justice,' 'bill,' 'democrats,' 'election,' 'media,' 'candidate,' 'vote,' 'speech,' 'deal,' 'Thursday,' 'voters,' 'congress,' 'abortion,' 'democratic,' 'Tuesday,' 'news,' 'racist,' 'white,' 'illegal,' 'presidential,' 'republicans,' 'supreme,' 'gay,' 'senate,' 'immigration,' and 'immigrants.'} This process was necessary to assess the model's performance for more abstract classes.

\paragraph{Performance Challenges and Critical Questions} The F1 scores attained in the realm of AC image classification reveal a significant challenge: they tend to be notably lower when compared to classes of a more concrete nature. Furthermore, the work that boasts the highest F1 score \cite{thomas2021predictingvisual} only marginally outperforms the second-best in this regard. This observation is intriguing, considering that the highest-performing work benefits from two distinct advantages over its counterparts. Firstly, it operates with a substantially larger dataset, comprising one million images, two orders of magnitude greater than those employed by other works. Secondly, this work leverages a  training approach that might be considered considered `cheating', as during the training phase, the model has access to and learns from textual documents (news articles) the images were collected from. We decided to keep this work as, even though this added textual information confers an advantage, at test time the approach does not have access to text.  Despite this substantial data magnitude and training advantage, the F1 scores achieved remain relatively low, illuminating the intricate and demanding nature of AC image classification tasks.

\paragraph{Importance of Hybrid Approaches} The prevalence of hybrid models in half of the surveyed works suggests that a purely statistical approach may be insufficient for AC image classification tasks. The inclusion of symbolic knowledge, whether through intermediary features or external knowledge and reasoning, appears to be a necessity in this domain. This observation highlights the growing relevance of hybrid AI systems for addressing the complexity of AC image classification.

\section{Relevant Datasets}
\label{section:survey_datasets}

Many popular image datasets used in CV offer low or no coverage of non-concrete concepts, such as ImageNet \cite{deng2009imagenet} and Tencent ML-Images dataset \cite{wu2019tencentmlimages}. Others, like JFT \cite{sun2017revisitingunreasonable}, an internal dataset at Google, include non-concrete classes but are not publicly released. We identified additional datasets that contain at least some images annotated with non-concrete labels.

\subsection{Natural Images}
\begin{itemize}
    \item \textbf{NUS-WIDE} \cite{chua2009nuswiderealworld}: Comprising 27,000 Flickr images, each associated with hashtag-like tags (approximately 4,000 unique tags).
    \item \textbf{Open Images} \cite{kuznetsova2020openimages}: Comprises 9 million images with annotations for 19,794 classes from JFT. But while some non-concrete classes like \textit{peace}, \textit{pollution}, and \textit{violence} are present, they are long-tailed and explicitly designated as non-trainable.
    \item \textbf{MultiSense} \cite{gella2019crosslingualvisual}: Comprising 9,504 images annotated with English, German, and Spanish verbs.
    \item \textbf{VerSe} \cite{gella2016unsupervisedvisual}: Containing 3,518 images annotated with one of 90 verbs.
    \item \textbf{UNSPLASH}\footnote{\url{https://github.com/unsplash/datasets}}: The complete version of this dataset contains over 4.8 million high-quality photographs accompanied by 5 million keywords including some ACs.
    \item \textbf{BabelPic} \cite{calabrese2020fatality}: Comprising 14,931 images tagged with 2,733 non-concrete synsets, created by cleaning the image-synset associations from the BabelNet Lexical Knowledge Base.
    \item \textbf{Persuasive Portraits of Politicians} \cite{joo2014visualpersuasion}: Comprising 1,124 images of politicians, each labeled with ground-truth persuasive intents of 9 types and syntactical features of 15 types.
    \item \textbf{Politics} \cite{thomas2021predictingvisual}: Comprising 1 million images tagged with left- or right-wing political bias; each image accompanied by the text (news articles) in which they were originally embedded.
    \item \textbf{BNID BreakingNews} \cite{lyu2019attend}: Comprising approximately 10,000 images sourced from breaking news events labeled with 77 different classes. More than half of these classes are abstract representations not directly related to objects (e.g., law, policy, G20).
\end{itemize}

\subsection{Art Images}
\begin{itemize}
    \item \textbf{WikiArt Emotions} \cite{Mohammad2018}: Comprising 4,000 pieces of art featuring annotations for the emotions evoked in the observer.
    \item \textbf{ARTemis} \cite{achlioptas2021artemisaffective}: This dataset includes 455,000 emotion attributions and explanations provided by humans for 80,000 artworks sourced from WikiArt, including visual similes, metaphors, and subjective references to personal experiences.
    \item \textbf{SemArt} \cite{garcia2018howread}: A multi-modal dataset designed for the semantic understanding of art. It contains fine art painting images, each associated with attributes and a textual artistic comment.
    \item \textbf{The Tate Gallery Collection}\footnote{\url{https://github.com/tategallery/collection}}: Tate Gallery's collection metadata featuring 70,000 artworks tagged with a taxonomy covering a wide spectrum of concrete to non-concrete concepts.
    \item \textbf{ArtPedia} \cite{stefanini2019artpedia}: Encompasses a collection of 2,930 paintings and 28,212 textual sentences that not only describe the visual content of the paintings but also provide additional contextual information.
\end{itemize}

\subsection{Advertisement Images}
\begin{itemize}
    \item \textbf{Ad Dataset} \cite{hussain2017automaticunderstanding}: Comprising 64,832 image ads with comprehensive annotations that cover various aspects, including the topic, sentiment, persuasive strategies, and symbolic references employed in the ads.
\end{itemize}

\subsection{Datasets for Connecting ACs to Cultural Images}

While these datasets encompass labels or classes extending beyond traditional concrete concepts, they may not consistently offer high-quality tags for ACs. One contributing factor is the ambiguity surrounding the origins of these tags, often aggregated from online sources. In our comprehensive examination of each dataset, we have discerned select datasets where images bear explicitly recorded AC labels, denoting human-provided annotations. Informed by insights gained from scrutinizing CV works dedicated to tasks analogous to AC image classification (Section \ref{section:survey_ac_ic}), we have directed our attention toward cultural images, potentially featuring superior AC tags. We provide details of these datasets in Table \ref{tab:ac_datasets}

\begin{table}[h]
\centering
\footnotesize
\begin{tabular}{p{2.5cm}|p{3cm}|p{1.7cm}|p{6cm}}
\hline
\textbf{Dataset Name} & \textbf{Image Type} & \textbf{Size} & \textbf{Primary Focus} \\ \hline
Ads Dataset & advertisements & 13,938 & Features ACs as 'symbolism' \\
Tate Gallery & visual artworks & 70,000 & Rich tag taxonomy that spans concrete tags and ACs \\ 
ARTemis & visual artworks & 80,000 & Utterances include ACs  \\ 
ArtPedia & visual artworks & 2,930 & Visual descriptions include ACs \\ 
\end{tabular}
\caption{Datasets with explicitly recorded AC labels}
\label{tab:ac_datasets}
\end{table}

\paragraph{Ads Dataset}This dataset is one of the very few datasets that have explicit, single ACs as tags for its images. The subset with AC tags is composed of 13,938 images and prominently features symbolism associated with ACs. The dataset provides a list of 221 symbols, each accompanied by bounding boxes. These symbols often represent common abstract ideas such as \textit{danger}, \textit{fun}, \textit{nature}, \textit{beauty}, \textit{death}, \textit{sex}, \textit{health}, and \textit{adventure}.

\paragraph{The Tate Gallery Collection}The gallery's collection metadata, consisting of 70,000 artworks, is publicly accessible through a GitHub repository. The dataset boasts a rich tag taxonomy that spans a wide spectrum of concepts. It covers both concrete (e.g., \textit{vacuum cleaner} and \textit{shoe}) and non-concrete (e.g., \textit{consumerism} and \textit{horror}) subjects under categories like ``universal concepts'', making it a valuable resource for exploring abstract and non-abstract concepts.

\paragraph{ARTemis}Although this dataset primarily focuses on emotions, encompassing 455,000 emotion attributions and explanations related to 80,000 artworks from WikiArt, it extends its scope to include ACs like \textit{freedom} and \textit{love}. Notably, the dataset authors have conducted an analysis to gauge the degree of abstract versus concrete language within ARTemis. To measure abstractness or concreteness, they employed the lexicon introduced by \cite{brysbaert2014concreteness}, which assigns a rating from 1 to 5 reflecting the concreteness of around 40,000 word lemmas. In this assessment, a randomly selected word from ARTemis received a concreteness rating of 2.81, as opposed to COCO, which received a rating of 3.55 (with a statistically significant p-value).

\paragraph{ArtPedia}Comprises 2,930 paintings and 28,212 textual sentences.  Out of these sentences, 9,173 are specifically dedicated to providing visual descriptions. Upon manual examination, it becomes evident that these visual descriptions include references to ACs, similar to what can be found in ARTemis. An example of a visual description is ``Mistress and servant, a \textit{power} relationship, maybe some deeper emotional bondage," demonstrating the dataset's potential to be curated to explore and better understand ACs within visual art.

\section{Discussion}
\label{section:survey_discussion}
In the realm of human visual sensemaking and understanding, the term `high level' characterizes complex, subjective, and abstract visual reasoning, albeit with a diverse focus on various semantic units. Section \ref{section:survey_tip} delves into multidisciplinary analysis of this ``tip of the iceberg'' of high-level semantics and categorizes them into clusters based on their inherent characteristics and attributes. We identified four clusters of high-level semantics. These include \textit{commonsense semantics}, covering more objective and widely accepted semantic elements such as actions, activities, events, relationships, and object purposes, \textit{emotional semantics}, covering aspects related to emotions, moods, emotional cues, and individualized affects; \textit{aesthetic semantics}, which evaluates global aesthetic attributes contributing to overall image judgments; and \textit{inductive interpretative semantics}, encompassing complex, subjective, and culturally encoded elements like ACs, symbols, and symbolical values.

Our comprehensive analysis of 38 publications from top-rated CV venues in Section \ref{section:survey_ahlvu} provides a structured categorization of high-level semantic understanding in CV, facilitating a comprehensive understanding of the field's landscape. We have organized these high-level visual reasoning clusters into five main categories. \textit{Situational analysis} focuses on actions, activities, roles, and object purposes, incorporating abstract reasoning, situation recognition, and event recognition. \textit{Visual sentiment analysis} explores how images elicit emotional responses, encompassing emotion detection across various image types.\textit{Aesthetic analysis} centers on detecting or predicting aesthetic value in images across different categories. \textit{Social signal processing} delves into the recognition of non-concrete social aspects and the study of group attributes in images. Finally, \textit{visual rhetorical analysis} concentrates on discerning intrinsic and implicit meanings in images, particularly in the context of visual persuasion and advertisement understanding. These distinct clusters collectively provide a comprehensive framework for understanding high-level visual reasoning in CV, revealing the diverse areas of research and their interconnectedness.

Several critical lessons and emerging patterns have come to light from this survey of CV high-level tasks. First, there is a notable inclusion of sociocultural elements within this research domain, with research increasingly focusing on sociocultural aspects like emotions, relationships, and social events. This shift reflects the growing recognition of the importance of societal contexts within the field. Additionally, the survey shows that researchers are diversifying their approach to image types. While natural photographs remain a significant focus, there is a clear expansion into cultural images, signaling an evolving interdisciplinary approach in CV. Furthermore, the survey highlights the prevalence of task-specific datasets, underlining the necessity of tailoring data to meet the unique demands of various research objectives. Moreover, there's a clear trajectory of research evolution in the field. The survey demonstrates substantial growth in publications, particularly post-2012 and post-2017. This growth aligns with the ascent of deep learning and transformative moments in CV. It signifies the field's increasing awareness of its capability to address the intricate world of abstract semantics in visual data.

The in-depth survey into works dealing with tasks analogous to Abstract Concept (AC) image classification (Section \ref{section:survey_ac_ic},) offers valuable insights and lessons. Notably, it becomes evident that amassing humongous amounts of data, even with datasets comprising one million images, does not necessarily guarantee high F1 scores in AC image classification. This observation challenges the prevailing notion that massive data alone can address the complexity of this task, highlighting the need for more sophisticated approaches. Furthermore, adding textual information to the training process bestows an advantage, yet the F1 scores achieved, even with this favorable setup, remain relatively low. This underscores the intricate and demanding nature of AC image classification, emphasizing the need for novel techniques beyond data augmentation.

Another important lesson from the survey is the recognition of mid-level features, such as objects and facial expressions, as potentially crucial elements in AC image classification. Understanding the significance of these intermediary features can guide the development of more effective models, improving the accuracy of classifying ACs in images. Moreover, the prevalence of hybrid approaches in the surveyed works suggests that a purely statistical approach may fall short in the realm of AC image classification. To tackle the complexities associated with ACs in images, the inclusion of symbolic knowledge, whether through intermediary features or external knowledge and reasoning, emerges as a necessity. This finding highlights the growing relevance of hybrid AI systems that can seamlessly integrate statistical and symbolic knowledge to address the multifaceted nature of AC image classification tasks.

\section{Conclusions}
\label{section:survey_conclusion}

There is a significant body of work focused on automating high-level visual understanding tasks to mirror the most complex cognitive processes and subjective sensemaking inherent to human visual perception. This survey has revealed that a noteworthy focal point within this research landscape pertains to the investigation of social and socio-cultural cues and signals. These works explore semantic units that closely align with what cognitive science classifies as ``abstract concepts," encompassing social and cultural values and ideologies, denoted by various terms such as ``symbols," ``intents," or ``abstract topics." These are intricately tied to the domain of visual rhetorics, and our comprehensive exploration of this territory has unearthed several pivotal insights with direct implications for the advancement of high-level visual understanding, especially within the realm of AC image classification.

Foremost among these insights is the recognition that, even when operating with substantial datasets, including those comprising millions of images, achieving high F1 scores in AC image classification remains a formidable and persistent challenge. This observation prompts a reconsideration of the notion that accumulating vast amounts of data alone serves as the primary panacea \cite{thomas2021predictingvisual}. Additionally, the incorporation of supplementary information, such as textual content, and the judicious consideration of mid-level features such as objects and facial expressions, emerges as a critical avenue for enhancing performance. Critically, the prevalence of hybrid models in AC image classification work underscores the insufficiency of exclusively relying on statistical methodologies. The imperative inclusion of symbolic knowledge, whether through intermediary features or external knowledge and reasoning, is demonstrated as an essential component in this domain. This trend accentuates the growing significance of hybrid AI systems, poised to tackle the intricate and multifaceted challenges inherent to AC image classification tasks.



\bibliographystyle{ACM-Reference-Format}
\bibliography{output}





\end{document}